\documentclass[letterpaper, 10 pt, conference]{ieeeconf}  

\IEEEoverridecommandlockouts                              

\usepackage{graphics} 
\usepackage{epsfig} 
\usepackage{mathptmx} 
\usepackage{times} 
\usepackage{amsmath} 
\usepackage{amssymb}  
\usepackage{comment} 
\usepackage{subcaption}
\usepackage{hyperref}
\usepackage[ruled,vlined,linesnumbered]{algorithm2e}  

\newtheorem{theorem}{Theorem}
\newtheorem{corollary}{Corollary}

\newtheorem{definition}{Definition}

\title{\LARGE \bf
    Deploying Physically and Epistemically Compliant Agents by Distilling Belief-Space Task and Motion Planners
}

\author{Phillip J.K. Christoffersen$^{1}$, Rachel Ma$^{1}$, Patrick Barragán$^{1}$, and Dylan Hadfield-Menell$^{1}$
\thanks{$^{1}$MIT CSAIL. Corresponding author email: philljkc@mit.edu}
}

\begin{document}

\maketitle
\thispagestyle{empty}
\pagestyle{empty}

\begin{abstract}
In this work, we train agents to act under uncertainty according to specified, belief-dependent rules. Deep reinforcement learning (RL) can in theory train agents to follow rules if encoded into well-designed rewards, but in practice, RL fails to perfectly optimize these rewards, leading to ineffective or noncompliant agent behavior. Belief-space planners by construction produce behavior in line with developer specification, but require hand-programming online planners and belief-state estimators, adding significant engineering complexity at scale. We combine the two: we plan offline to generate a dataset of sample trajectories and then distill the plans into a policy that acts on raw observations. We propose two changes to planner distillation that address problems specific to history-based policies: Dual Averaging, a loss that combines per-timestep predictive loss and average loss on task changes, and $\rho$-Optimal Reinforcement, a feedback mechanism that reinforces the policy's own successful rollouts when these are near-optimal. We provide a sufficient condition under which static search problems with noiseless observation models is complete, and show that maximum-likelihood-observation planning implies this condition in static search with continuous beliefs. We create a challenging new domain, Scavenger, where an agent needs to gain information in order to succeed. We evaluate our distilled policy in two versions of Scavenger, one where information is provided visually and one where it is provided as a command in natural language. In either case, the learned policy attains a success rate above 90\% in Scavenger, compared with at most 5\% for RL. We also find that distilled planners transfer to settings unseen in training, with greater than 60\% success at following unseen and intentionally vague commands and greater than 75\% success at solving floorplans with more rooms than those seen in training.
    
\end{abstract}

\section{INTRODUCTION}

\begin{figure*}[t]
    \centering
    \includegraphics[width=\linewidth]{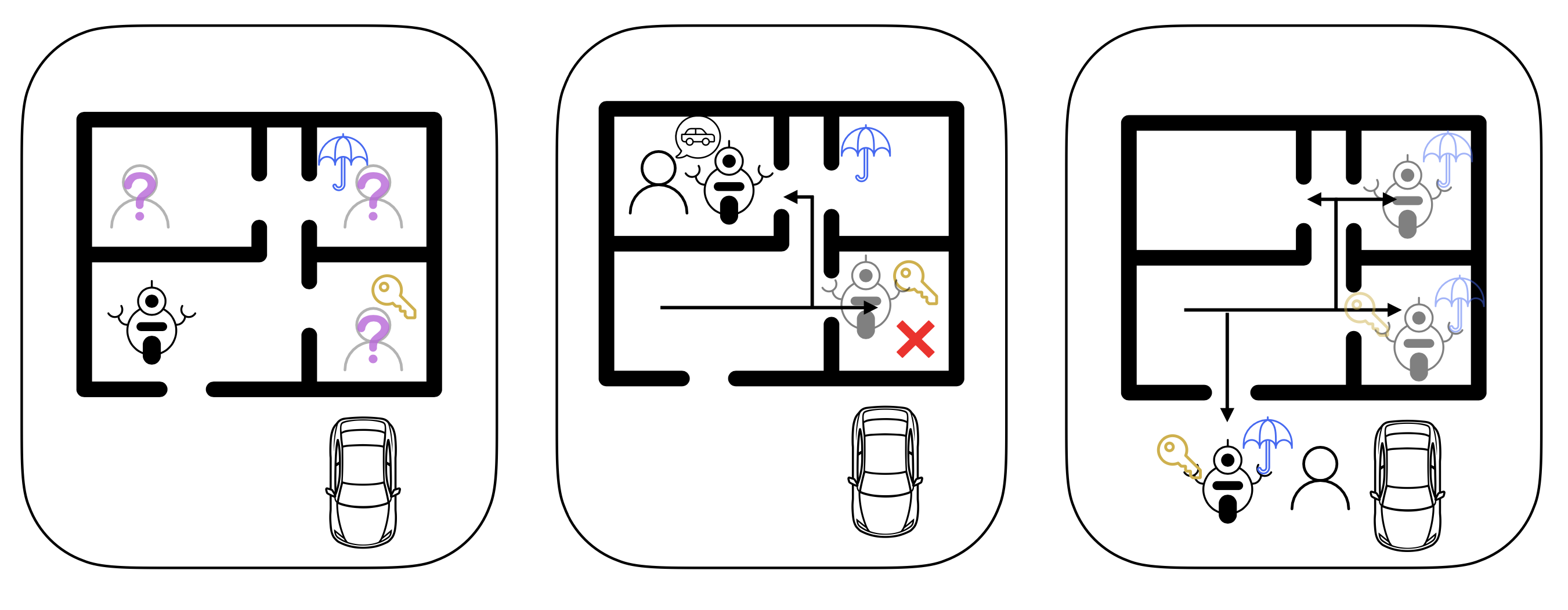}
    \caption{A worked example, where the robot must obey the epistemic rule: ``ensure the user leaves their home, and has what they need when they leave'', while being physically safe. The robot should be confident that the user has what they need before they leave the house. \textbf{Left:} The robot must first search for the user in the house, who may be in one of three locations. \textbf{Middle:} After finding the user, the robot asks them about what they might need for the day. \textbf{Right:} The robot hears the user is planning to go on a drive to see friends on a rainy day, and then brings them their keys and umbrella.}
  \label{fig:banner}
\end{figure*}

Our goal is to design capable AI agents that act under uncertainty, implementing behavior governed by specifications that depend on the agent's belief. Reinforcement learning (RL) often runs into challenges for such tasks, as rewards that fully capture a specification are complex, leading to optimization difficulties and convergence to suboptimal policies. As a result, intended behavior does not reliably correspond to RL policies' learned behaviors. On the other hand, online belief-space planners produce behavior that complies with specifications by construction. But implementing planners often has significant engineering overhead, particularly under partial observability, requiring online belief estimation from raw sensor data, and low-latency planning given the need for repeated replanning. These are difficult to implement efficiently for general-purpose agents. We extend prior work~\cite{mcdonald2022guided} on planner imitation to train capable and efficient policies for long-horizon problems with partial observability, planning offline with access to a simulator and distilling those plans into a policy that acts directly on raw observations.

For example, consider a robot that needs to help a person leave their house in the morning. We want it to search the home to find the person, listen to what they need and decide how best to help, for example by finding their car keys and umbrella if they plan on a day trip. All of this needs to happen while the robot complies with physical safety constraints, such as avoiding obstacles, and epistemic constraints, such as ensuring that the user doesn't leave before they have everything they need for the day. Reinforcement learning struggles with multi-step, information-dependent tasks like this, where exploration and credit assignment are challenging. Furthermore, even when RL solves a task, it may not give us the behavior we want. For example, if other observations predict when the umbrella is needed, nothing in the reward requires the robot to ask, and the learned policy might skip actually finding the human.

A belief-space planner, on the other hand, can avoid these issues and solve long-horizon tasks with epistemic constraints. However, online belief estimation makes these planners difficult to deploy: an online planner must estimate a full model of the environment in which the agent operates in order to produce full plans. Planner imitation addresses this deployment problem by training policies to imitate a planning system with access to simulator information not seen online. This way, the agent only needs to identify features from its observations which are necessary for predicting tasks and controls. These methods are difficult to apply to belief-space planning, because the dependence on the history of observations makes the problem of distribution shift from a learned policy more severe. Existing feedback mechanisms~\cite{mcdonald2022guided} focus on low-level control, which is relatively simple in this setting, give high-level feedback only when the policy takes an impossible action, and cannot practically replan from every state the policy reaches. 

Concretely, we make the following contributions:

\begin{itemize}
    \item We make two algorithmic improvements that enable planner imitation with history-based policies: a new loss that upweights task changes in observation-action histories, and a new, simple feedback mechanism that reinforces successful policy rollouts with low regret against planned behavior and provides planner demonstrations otherwise.
    \item We find sufficient conditions under which belief-space task and motion planning is \textit{complete} under noise-free observations for static search. We use this to show that planners assuming maximum-likelihood-observations are complete for continuous priors $B_0$. 
    \item We demonstrate our method in Scavenger, a novel domain with irreversible actions and natural language commands. Our method attains a success rate above $90\%$, compared to at most $5\%$ success for reward-engineered RL, and transfers to unseen vague commands and floorplans with up to three times as many rooms as seen in training.
\end{itemize}

We implement this work in OpenPLAN, our code base for belief-space planning and policy distillation, which we release with this paper. More broadly, this work supports the development of safe AI by laying the groundwork for training systems that satisfy epistemic constraints: in Scavenger, the policy learns to hold an irreversible action until it has the necessary information, without an explicit belief estimator.

\section{RELATED WORK}

\subsection{RL in POMDPs}

Several works propose learning compact belief representations that enable tractable reinforcement learning. Gangwani et al.~\cite{gangwani2020learning} show that latent belief embeddings can be learned from demonstrations, allowing policies to generalize beyond explicit filtering. Similarly, Wang et al.~\cite{wang2023learning} use recurrent and latent-variable models to capture hidden state uncertainty directly from data, making deep RL more effective in partially observable settings. 
Kim et al.~\cite{kim2023belief} introduce a method that projects high-dimensional belief distributions into lower-dimensional subspaces, enabling efficient policy learning in environments with delayed rewards and feedback. The drawback with these techniques is that they address the core issues only indirectly: they either require auxiliary datasets or demonstrations that are informative enough to learn identifiable belief states for tasks, or must explore sufficiently well to generate a good data set for identifying the belief state. We avoid both problems by allocating the job of state estimation and belief-tracking to the simulator offline, and then imitating the planner behavior, providing additional supervision on policy failure to accomplish the task.

\subsection{Belief-Space Task and Motion Planning}

Belief-space task and motion planning has a long tradition in the planning literature \cite{platt2010belief, bonet2011planning, kaelbling2013integrated, hadfield2015modular, garrett2020online}. Our planner draws on several of these. We use the execution-monitoring schema from \cite{kaelbling2013integrated} as a semantics for generating belief-space plans. We use the particle representation of \cite{hadfield2015modular}, updating the belief with a particle filter, and predict observations from the most likely particle; this requires only sampling and black-box access to the observation model. Platt et al.~\cite{platt2010belief} plan assuming maximum-likelihood observations. Under Gaussian beliefs and noise, that is the observation the mean state would produce. In noiseless search, however, the \texttt{null} observation carries a point mass, so the maximum-likelihood observation is typically \texttt{null}, and a planner that assumes it never expects to find what it is searching for. We therefore predict observations from the most likely state while still planning toward belief goals. While we focus on world-state uncertainty, it is possible for belief-space planning to handle many more kinds of uncertainty beyond those we consider, including physical or SLAM uncertainty \cite{curtis2024partially}. Our framework uses the Fast Downward \cite{helmert2006fast} planner for task planning, and solves motion planning instances using Sequential Quadratic Programming \cite{hadfield2016sequential}. 

\subsection{Imitating TAMP Solvers}

We differ from the above planning methodologies by using the planner \textit{as a source of data}. The final output from our architecture is a policy that is trained on rollouts in a separate simulator from the planner. This is advantageous because (1) it allows the system designer more flexibility in the space of abstractions to plan over, potentially lessening the burden on the planner and the system designer in formalizing TAMP instances, and (2) it allows a policy to be trained on generic observations, potentially different from the space planned over, allowing scaling to much more complex environments (e.g., generating a policy that operates with image input, by imitating plans which operate over much simpler abstractions). To implement this, we extend the work of \cite{mcdonald2022guided}, which applied this framework to fully observed domains, to allow for uncertain quantities, and implement a framework for an agent to take observations and maintain its internal belief of uncertain quantities. 

Several other works have created analogous workflows to this, in the fully observed domain, e.g., \cite{driess2021learning, dalal2023imitating, sahoo2026tandemtaskmotionplanning}. Paxton et al.~\cite{paxton2017combining} trained policies that are similar to TAMP solvers through combining Monte-Carlo Tree Search with deep Q-learning. Kase et al.~\cite{kase2020transferable} trained hierarchical policies for TAMP from visual inputs. However, working in the fully observed case sidesteps significant optimization and engineering complexities for distillation. For example, working under partial observability requires policies that consume the entire observation-action history to produce decisions, which have much more brittle training dynamics, being more sensitive to variable history lengths and state visitation distribution shifts from feedback. To adapt to this more complex setting, we find that upweighting timesteps with task changes, and reinforcing successful policy trajectories, are more robust for training history-based policies than hand-engineered feedback.

\section{PRELIMINARIES}

\subsection{POMDPs and the Belief-Space Transform}

POMDPs form the standard model for decision-making problems under uncertainty in RL and in robotics broadly.  

\begin{definition}[POMDPs]
    A POMDP (partially observable Markov decision process) is an 8-tuple $$P = \langle S, A, S_0, T, R, \gamma, O, \mathcal{O} \rangle$$ where $S$ is a state space, $A$ is an action space, $S_0 \in \Delta S$ is an initial state distribution, $T: S \times A \to \Delta S$ is a transition function, $R: S \times A \to \mathbb{R}$ is a reward function, $\gamma \in [0, 1)$ is a discount factor, $O$ is an observation space, and $\mathcal{O}: S \to \Delta O$ is an observation model.
\end{definition}

The objective in a POMDP is to find a policy maximizing expected discounted return given an observation-action history. Solving POMDPs poses serious computational challenges because the agent, per-timestep, only perceives an observation $o \in O$ of the true state and not the true state $s \in S$. It thus needs to compute the posterior probability of each $s \in S$, conditioned on its observation-action history $o_1, a_1, \ldots, o_t$, and filtered through transition function $T$. The space of distributions $\Delta S$ is called belief space, and we can transform a POMDP into an MDP, called the belief MDP, whose transitions update the current belief $b$ after action $a$ and observation $o$ by the Bayesian update 
$$b'(s') \propto \mathcal{O}(o \mid s') \int_s T(s' \mid s, a)\, b(s)$$

as per \cite{kaelbling1998planning}. We are interested in training agents that satisfy constraints defined on belief states $\Delta S$. 

\subsection{Task and Motion Planning}

We now outline task and motion planning (TAMP) as a general problem class, providing a hierarchical model for MDPs (or POMDPs via the belief-space transform). In TAMP, a system designer provides a set of high-level actions $\mathcal{A}$ called operators. Formally: 

\begin{definition}[Task and Motion Planning]
    A task and motion planning domain can be written as:
    $$\mathcal{T} = \langle X, U, \mathcal{A} \rangle$$ such that $X$ is a state-space of the world, $U$ is a control space, and $\mathcal{A}$ are maps $a: X \to \mathcal{P}(X^*)$, i.e., from each initial state $x$ to a set of admissible state histories $\tau^x \in \mathcal{P}(X^*)$ for motions under the definition of that operator (including the empty set if that operator cannot be applied at $x$). 
\end{definition}

The objective of task and motion planning is to produce, given an initial state $x_0 \in X$ and goal subset $G \subset X$, a trajectory of operators $\langle a_1, \ldots, a_T\rangle$ and a corresponding sequence of controls $\langle \tau^u_1, \ldots, \tau^u_T \rangle$ with lengths $t_1, \ldots, t_T$ satisfying operator constraints, and reaching $G$. Formally, this means every induced trajectory of states $\tau^x_i$ satisfies

$$\tau^x_i \in a_i(x_{\sum_{0 \le j < i} t_j})$$ 

and further that $x_0$ is the initial state in $\tau_1^x$, and the final state of $\tau_T^x$ is an element of goal subset $G$. Solving these problems has a discrete combinatorial component in choosing the $a_i$ and then continuous components in choosing appropriate $\tau^u_i$. Typically, a symbolic planner (such as Fast Downward \cite{helmert2006fast}) uses a logical representation of the operators to determine what sequence of operators to attempt, and then a motion planner (typically a sampler or optimizer) will attempt to produce a corresponding $\tau$ for each $a$ in sequence. 

\subsection{Belief-Space Task and Motion Planning}

We implemented the belief-space planning semantics of \cite{kaelbling2013integrated}, specifically as modeled in \cite{hadfield2015modular}. The main difference between belief-space TAMP and standard TAMP is that an agent cannot perfectly predict the belief-state effects of its actions. For example, when carrying out a search task, an agent cannot know for certain whether or not it will actually observe an object at a point. Belief-space planning gets around this by planning based on assumptions about what it might observe and execution monitoring when running a plan to see if this was what actually was observed, replanning if necessary. \autoref{alg:plan} outlines our planner.

\subsection{Guided Imitation of Task and Motion Planning \cite{mcdonald2022guided}}

In order to distill the belief-space planning semantics from earlier, we build on the distillation pipeline provided in \cite{mcdonald2022guided}. In that work, the distilled policy is hierarchical, consisting of a high-level network $\pi_H$ predicting the current task planner action in each timestep, and a set of low-level control policies $\pi_L$ predicting controls for the robot to enact in each timestep. They provide a distributed implementation, where generation of data for $\pi_H, \pi_L$, training $\pi_H, \pi_L$, and generating rollouts to evaluate the policy. They leverage these rollouts to provide policy-aware feedback. They give feedback to $\pi_L$ by biasing the motion planner to prefer nearby states to rollouts. They also give feedback to $\pi_H$ by altering the distribution of problems the planner solves, in particular to prevent the policy from taking illegal actions. This second feedback mechanism, called the Task Exploration Node, is complicated to implement, and shifts the distribution of visited states considerably. We sidestep these complexities with our simplified feedback mechanism, as outlined in \autoref{sec:methods_rho}. 

\begin{algorithm}
\DontPrintSemicolon
\SetKwProg{Fn}{def}{:}{}
\SetKw{KwRaise}{raise}
\caption{Belief-Space TAMP}
\label{alg:plan}
\KwIn{initial state $s_0$, planning domain $\mathcal{D}$}
\Fn{\textnormal{\texttt{plan}}$(s_0, \mathcal{D})$}{
  $s \leftarrow s_0$\;
  $i = 0$\;
  $\mathit{solved} \leftarrow \textsc{False}$\;
  $x \leftarrow \texttt{parse}(s)$\;
  \While{\KwSty{not} $\mathit{solved}$}{
    \tcc{task planning}
    $\langle a_i,\dots,a_n \rangle,\ \mathit{ok} \leftarrow \texttt{symb\_plan}(x, \mathcal{D})$\;
    \lIf{\KwSty{not} $\mathit{ok}$}{\KwRaise ``Symbolic plan failed''}
    \tcc{trajectory refinement}
    $\langle \tau^u_i,\dots,\tau^u_n \rangle,\ \mathit{ok} \leftarrow \texttt{refine}(\langle a_i,\dots,a_n \rangle, \mathcal{D}.samplers)$\;
    \lIf{\KwSty{not} $\mathit{ok}$}{\KwRaise ``Refinement failed''}
    \tcc{plan rollout}
    $x_{\mathrm{fail}} \leftarrow \texttt{rollout\_from\_state}(\langle \tau^u_i,\dots,\tau^u_n \rangle, x)$\;
    \eIf{$x_{\mathrm{fail}} = \varnothing$}{
      $\mathit{solved} \leftarrow \textsc{True}$\;
    }{
      $x \leftarrow x_{\mathrm{fail}}$ \tcp*{Replan from failure}
      $i \leftarrow \texttt{fail\_idx}$\;
    }
  }
  \KwRet $(\langle a_1, \tau^u_1 \rangle, \ldots, \langle a_T, \tau^u_T \rangle)$\;
}
\end{algorithm}

\section{METHOD}
\label{sec:methods}

In this section, we outline two improvements to \cite{mcdonald2022guided} that adapt it to training history-dependent policies for POMDPs. The crucial difficulty to solve in distilling planners is providing a training algorithm and network architecture amenable to predicting the next task as a discrete prediction, and the next motion as a continuous prediction. We adapt their hierarchical policy architecture, using separate $\pi_H$ and $\pi_L$ for action and control policies, but make two changes to the training algorithm. Dual Averaging pushes history-based policies to track the action and control timescales separately in a stable way. $\rho$-Optimal Reinforcement reinforces successful, near-optimal trajectories, providing feedback to the policy without significantly altering the distribution of histories seen in the data buffer through training. 

\subsection{Dual Averaging}
\label{sec:methods_loss}

We modify the loss function for $\pi_H$ under partial observability. The prior work \cite{mcdonald2022guided} used simple standard cross-entropy losses and behavior cloning for action prediction. In our domain, trajectories vary significantly in the number of different actions that $\pi_H$ will take, and the number of controls executed by $\pi_L$ in each high-level action. This difficulty is compounded by the dependence of predictions on the full history of observations and actions, which makes $\pi_H$ predictions autocorrelated. This means $\pi_H$ will default to continuing the current action and, correspondingly, make predictions biased against altering its current action. We correct for these issues by designing a loss $\mathcal{L}_{dual}$ for $\pi_H$, for each trajectory $\tau$, which takes the average loss over all timesteps and combines it with an loss averaged over timesteps where the task planner changes actions. In other words, given a base loss $\mathcal{L}$, we have the dual averaging loss $\mathcal{L}_{dual}$ over trajectory $\tau$ is:

$$\mathcal{L}_{dual}(\tau) = \alpha \left(\frac{\mathcal{L}(\tau)}{|\tau|}\right) + \beta \left(\frac{\mathcal{L}(\tau[t: a[t] \ne a[t-1]])}{|\{t: a[t] \ne a[t-1]\}|}\right)$$

We average these per-trajectory to account for variable length plans with different numbers of task changes.

\subsection{$\rho$-Optimal Reinforcement}
\label{sec:methods_rho}

We modify the Task Exploration Node feedback mechanism from \cite{mcdonald2022guided} for $\pi_H$. Since we consider learning in partially observed tasks, $\pi_H$ needs to consider the full observation-action history when making predictions, since information from early in the history might affect which course of action to take (e.g., which room in a home to search next depends on all rooms seen so far). Learning, therefore, needs some on-policy feedback to succeed. The feedback mechanism of~\cite{mcdonald2022guided} corrects for illegal high-level actions (e.g., actions with violated pre-conditions). However, in many partially observed problems, most task-policy failures are legal, but suboptimal (e.g., the robot can continue to search however long it might want, but it may not be effective to do so). 

We found that a simple strategy of \textit{reinforcing policy successes} worked well. Instead of directly detecting suboptimal actions, we compare a successful rollout to a successful planned trajectory. If the policy's trajectory comes within a factor $\rho \ge 1$ of the length of the planner's, we include it as training data. Otherwise, we push the planner's trajectory (if the planner produces a valid plan). $\rho > 1$ allows the policy to reinforce successful behavior that the policy has demonstrated it can enact, without reinforcing actions that are overly suboptimal.  Some suboptimality helps because it gives the learner more flexibility. This is outlined in \autoref{alg:feedback}.

\begin{algorithm}
\DontPrintSemicolon
\SetKwProg{Fn}{def}{:}{}
\caption{$\rho$-Optimal Reinforcement}
\label{alg:feedback}
\KwIn{optimality ratio $\rho$, planning domain $\mathcal{D}$, buffer $\mathcal{B}$, policy $\pi$}
\Fn{\textnormal{\texttt{feedback\_decision}}$(\rho, \mathcal{D}, \mathcal{B}, \pi)$}{
  $\tau^{\pi} \leftarrow \texttt{fetch\_policy\_rollout}(\pi)$\;
  $\tau^{F} \leftarrow \texttt{plan}(\tau^{\pi}.\mathit{init}, \mathcal{D})$ \tcp*{Alg.~\ref{alg:plan}}
  \uIf{$\tau^{\pi}.\texttt{is\_success}$ \KwSty{ and } $\left(\neg\,\tau^{F}.\texttt{is\_success}
       \textbf{ or } \texttt{len}(\tau^{F}) \leq \rho \cdot \texttt{len}(\tau^{\pi})\right)$}{
    $\mathcal{B}.\texttt{push}(\tau^{\pi})$\;
  }
  \ElseIf{$\tau^{F}.\texttt{is\_success}$}{
    $\mathcal{B}.\texttt{push}(\tau^{F})$\;
  }
}
\end{algorithm}

\section{THEORY}

Since a belief-space planner relies on replanning in order to solve problems, it is possible that planning never terminates, or takes many replans (i.e., iterations of the outer loop of \autoref{alg:plan}), in order to find a plan. Worse still, optimism can lead a planner to get stuck and never find a valid plan. However, many robotic problems under uncertainty take the form of static search for which we show belief-space planning \textbf{is} complete.  

\begin{theorem}
    Consider a POMDP with state space $S = S_{obs} \times S_{unobs}$, where $S_{obs}$ is fully observed and no action changes $S_{unobs}$. Assume:

    \begin{itemize}        
        \item (Noiseless observation model) $\mathcal{O}$ is an observation model of form 
        
        $$\mathcal{O}(s) = \begin{cases}
            s_{unobs} & \text{if } s_{unobs} \in A(s_{obs}) \\
            \texttt{null} & \text{otherwise} 
        \end{cases}$$
        
        for some set map $A: S_{obs} \to \mathcal{P}(S_{unobs})$
        
        \item ($\delta$-prior-efficient planner) The planner can always replan from an arbitrary state in $S_{obs}$ and belief in $\Delta S_{unobs}$, and whenever the planner replans from a belief state $B$ which is uncertain about $S_{unobs}$, the planner produces trajectories $\tau^x_i$ for which $\bigcup_{x \in \tau^x_i}A(x_{obs})$ contains a set of $B_0$-probability at least $\delta$ within the support of $B$.
    \end{itemize}

    Then, the maximum number of replans before an agent becomes certain of $S_{unobs}$ is $O(\frac{1}{\delta})$ as $\delta \to 0$. 
\end{theorem}

\begin{proof}
    Consider what happens to the robot belief, in any rollout. If $s_{unobs} \in \bigcup_{x \in \tau^x_i}A(x_{obs})$ at any point, then the robot receives observation  $s_{unobs}$ along its trajectory and at that action becomes certain of $s_{unobs}$. Hence, if the robot is uncertain about $S_{unobs}$ after replan $i$, it must be that $s_{unobs} \notin \bigcup_{x \in \tau^x_j}A(x_{obs})$ at any prior $0 \le j \le i$. 
    
    Thus, at replan iteration $i$, the planner's belief $B_i$ exactly tracks candidates for $s_{unobs}$ that have not been observed to this point. So, $\delta$-prior-efficiency gives that at each replan round, either a $B_0$-probability $\delta$ set of previously unexplored possibilities for $S_{unobs}$ is ruled out or the planner becomes certain of the true $s_{unobs}$. Hence, the maximum number of possible replans before the planner becomes certain of $s_{unobs}$ is $\lceil \frac{1}{\delta} \rceil + 1$ (in the worst case, it could have a $B_0$-probability-$0$ subset to explore if $\delta = \frac{1}{N}$), asymptotically $O(\frac{1}{\delta})$ as $\delta \to 0$. 
\end{proof}

\begin{corollary}
    In static search problems with noiseless observation models, $\delta$-prior-efficient planners are complete. 
\end{corollary}

\begin{proof}
    Since we have a bound on the number of replans, we have shown $\delta$-prior-efficient planners terminate. Hence, the planner will always, in finite time, resolve their uncertainty.
\end{proof}

\begin{corollary}
    In static search problems with noiseless observation models and continuous priors $B_0$ (i.e. for which $\mathbb{P}_{B_0}[s=s_{unobs}] = 0$ for any $s_{unobs}$) if maximum-likelihood-observation planners can produce valid replans from any initial $S_{obs} \times \Delta S_{unobs}$, they are complete. 
\end{corollary}

\begin{proof}
    First, we show that maximum-likelihood-observation planners which can always replan are $\delta$-prior-efficient, for $\delta=1$. To see this, it is enough to show that the initial proposed plan must cover a probability-one subset of $B_0$. For contradiction, assume otherwise: we have a proposed initial plan with $\bigcup_{x \in \tau^x_i}A(x_{obs})$ being probability $\delta < 1$ of the prior $B_0$, when replanning from an uncertain $B$, and which does not resolve the planner uncertainty. By assumption, the likelihood under $B$, a continuous distribution on a subset of $B_0$, of observing any given $s_{unobs}$ is $0$. Since $> 1 - \delta > 0$ of the prior is unvisited by this replan, the likelihood of observing \texttt{null} exactly along the replan would be strictly positive. So a replan covering less than an almost sure subset of $B_0$ would be guaranteed to never assume any observation of an $s_{unobs}$. But this is for contradiction: the planner has not attained its goal under assumed observations, hence this replan is invalid. So it is $\delta$-prior-efficient for positive delta, and hence, it is complete. 
\end{proof}

Notably, $\delta$-prior-efficiency implicitly assumes completeness for fully-observed problems. This corresponds to action irreversibility for search-relevant actions: if attempting to resolve uncertainty for part of your prior leads to it becoming impossible to explore other parts of the domain later, it would not be possible to do $\delta$-prior-efficient planning from that point on, since eventually you could not guarantee that inaccessible parts of the prior remain explored. These assumptions, however, still model a broad set of robotic domains and TAMP planners: POMDPs without modeled sensor noise and tasked with resolving uncertainty about uncertain quantities, and when planning assuming maximum-likelihood observations. $\delta$-prior-efficiency assumes that the planner, when replanning from any belief state $B$, is guaranteed to either (1) resolve its uncertainty or (2) plan for a set of points which observes a $B_0$-probability $\delta$ set of candidate values for $s_{unobs}$. This gives a novel definition of efficiency for the replanning logic in belief-space planning. This definition assisted us in the design of our planner: in \autoref{sec:experiments}, we developed a planner assuming the maximum-likelihood-observation in the most-likely-state. This allowed us to propose shorter plans than standard maximum-likelihood-observations per replan, at the cost of potentially more replans. Since this theory gives a bound on the maximal number of replans, having checked $\delta$-prior-efficiency (the planner in \autoref{sec:experiments} satisfies this for $\delta=\frac{1}{6}$), this ensured the planning cost would not grow excessively. Broadening this definition to a greater POMDP class is a question of interest for the theoretical foundations of robotics.

\section{EXPERIMENTS}
\label{sec:experiments}

We compare our method with flat and hierarchical variants of Proximal Policy Optimization (PPO) \cite{schulman2017proximal}, as well as the prior work our method builds on \cite{mcdonald2022guided}, to check whether the distilled planning pipeline attains a higher rate of compliance with complex specifications. We also perform ablations to show the impact of $\rho$-Optimal Reinforcement and Dual Averaging on distilling planners. We perform our experiments in Scavenger, a new long-horizon, partially observed domain. 

\begin{figure*}[t]
  \centering
  \begin{subfigure}[height=3cm]{0.33\textwidth}
    \includegraphics[width=\linewidth]{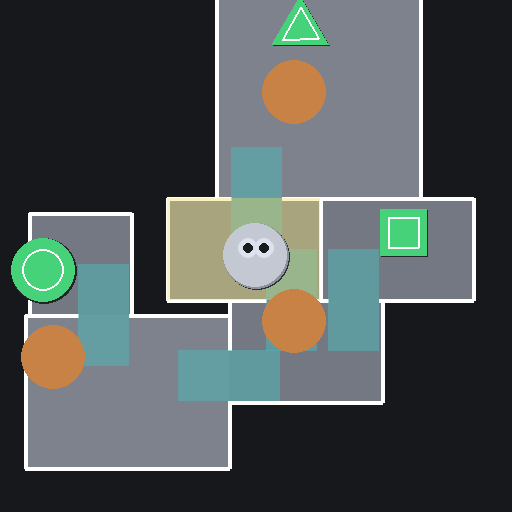}
    \caption{A sample initial condition }
    \label{fig:banner-a}
  \end{subfigure}\hfill
  \begin{subfigure}[height=3cm]{0.33\textwidth}
    \includegraphics[width=\linewidth]{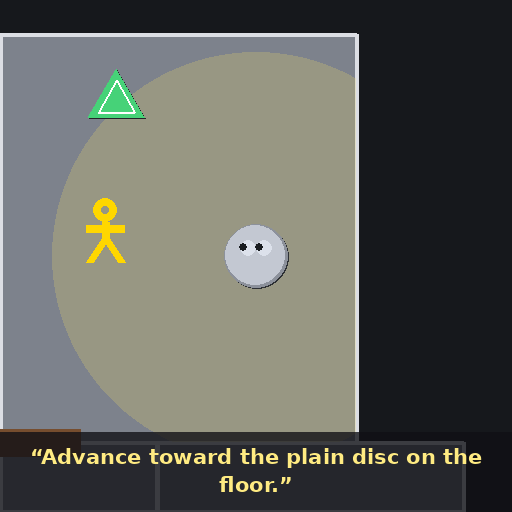}
    \caption{The user language hint in (a)}
  \end{subfigure}\hfill
  \begin{subfigure}[height=3cm]{0.33\textwidth}
    \includegraphics[width=\linewidth]{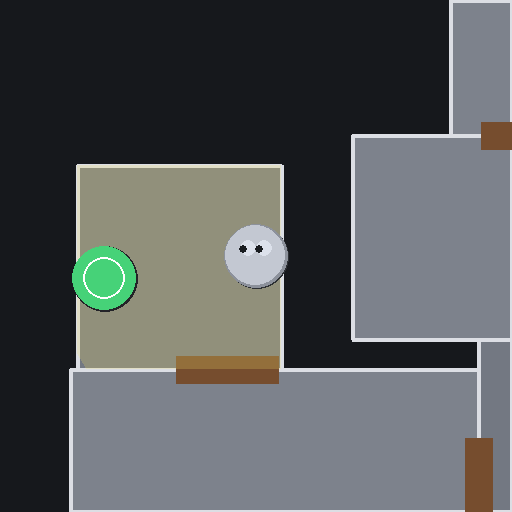}
    \caption{``Guessing'' a marker}
  \end{subfigure}
  \caption{An overview of the Scavenger domain. \textbf{(a)} The agent starts in a random room for a procedurally generated floorplan, with a set of visual markers, a geometry visualizing connectivity between rooms, and a visual prior denoting a set of possible locations. \textbf{(b)} The robot must search among possible locations for a hint, either represented visually or as a command in natural language. \textbf{(c)} It succeeds when it approaches the marker hinted at, and then takes a relevant ``guess'' action when it is sufficiently close to that marker. Guessing before it is certain of the correct marker, or guessing the wrong marker, counts as a failure.}
  \label{fig:Scavenger_domain}
\end{figure*}

\subsection{Planning In The Scavenger Domain}

To evaluate our ability to train agents balancing goal pursuit with physical and epistemic compliance, we implement Scavenger, modeled on the childhood game of a scavenger hunt. In this domain, an agent is randomly placed in a procedurally generated multiroom floorplan in which a set of shape markers, and markers denoting possible locations for a hint, are visible. It can observe a zoomed-in view, which shows the room it is in and a zoomed-out ``minimap'' view showing more of the floorplan, both 128$\times$128 RGB images. It can navigate with x- and y-position control, at most 0.3 units at a time. In order to succeed at this task, the agent must locate the hint and then ``guess'' by approaching the relevant marker and taking a guess action. The agent must never hit a wall while navigating and not guess early nor guess the wrong marker, in order to succeed. We experimented with two variants of the domain, a \textbf{Visual Hint} variant where the hint was visually rendered as the desired shape in a different color and a \textbf{Language Hint} variant where it was given in the form of a direct language command, rephrased 100 times per marker by Claude Opus. The floorplans vary between three and six possible rooms, each room being axis-aligned and between two and four units in width. Each floorplan has three floor markers: a circle, a square, and a triangle. Details for the TAMP formalization of Scavenger can be found in \autoref{section:formalism}. Notably, we do not use maximum-likelihood observation to implement the planner, but instead leverage a maximum-likelihood observation from the most-likely state, in order to shorten plan length (maximum-likelihood-observations, for static search tasks, plan a search path hitting every possible location).

\subsection{Experimental Setup}

\begin{figure*}[t]
  \centering
  \begin{subfigure}{0.49\textwidth}
    \includegraphics[width=\linewidth]{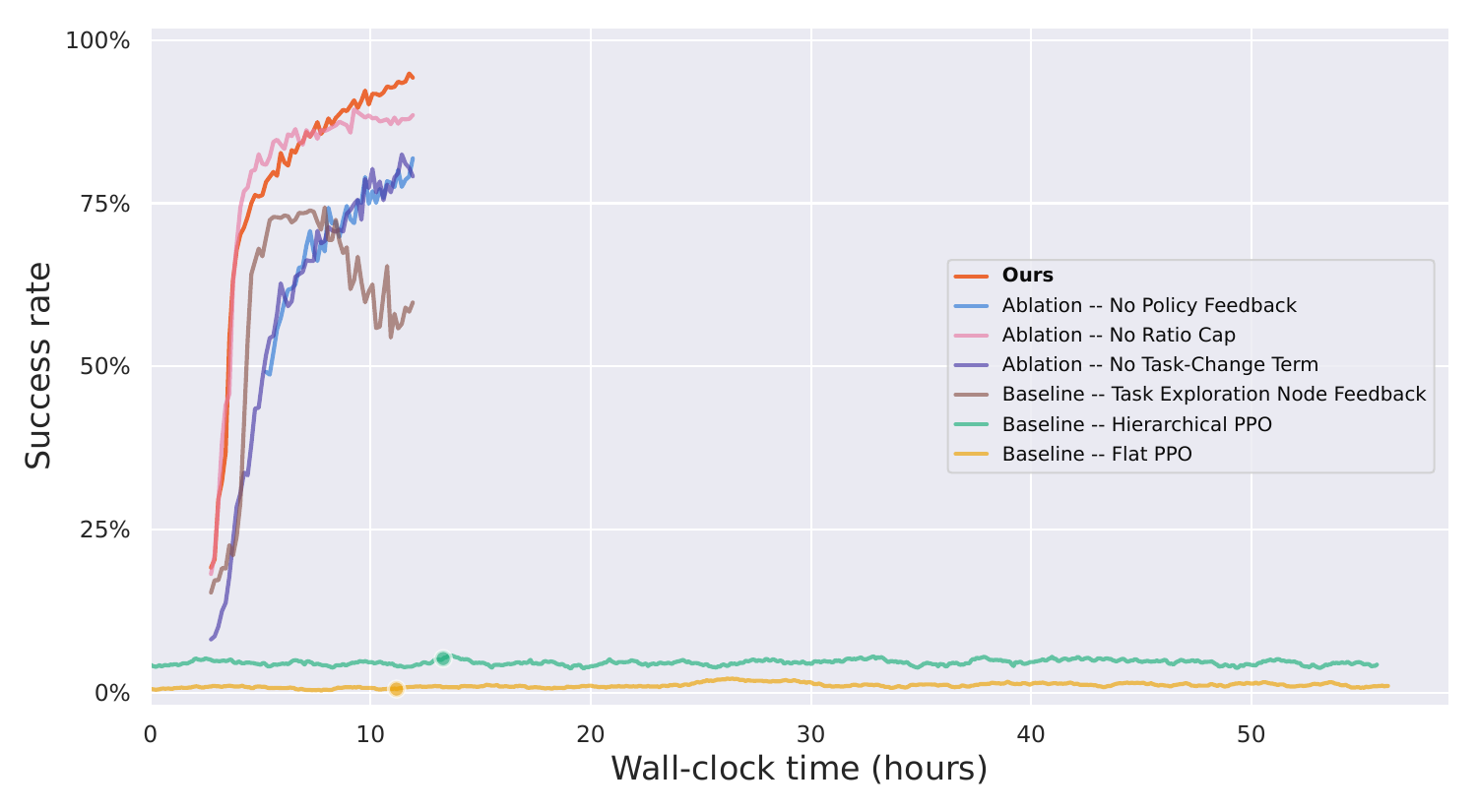}
    \caption{Scavenger with Visual Hints}
  \end{subfigure}\hfill
  \begin{subfigure}{0.49\textwidth}
    \includegraphics[width=\linewidth]{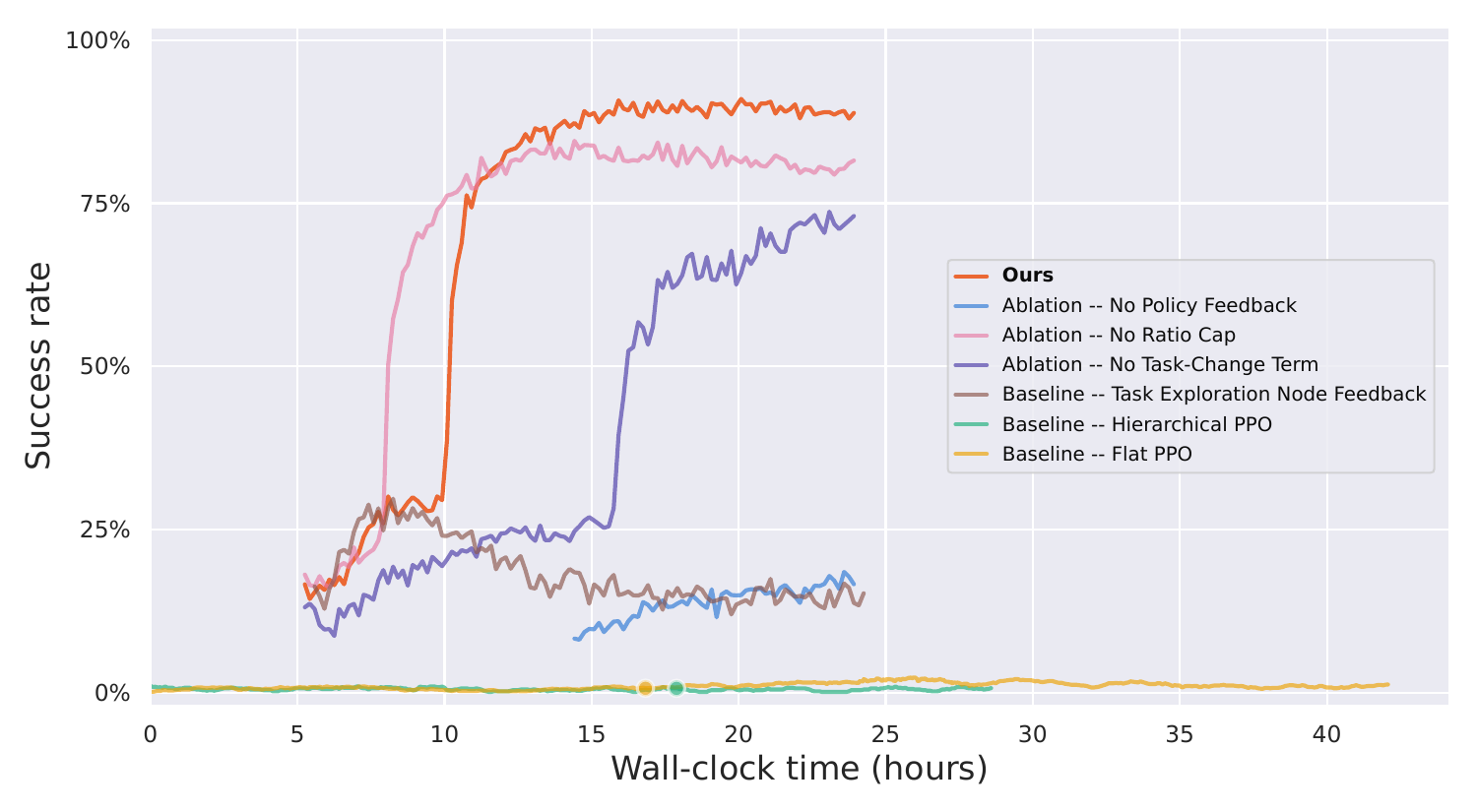}
    \caption{Scavenger with Language Hints}
    \label{fig:results_lang}
  \end{subfigure}
  \caption{Success rates vs.\ wall-clock time on Scavenger. The dots on the baseline runs refer to the point in baseline runs where as many simulation timesteps are used for training the policies as our method. In all cases, our method attains at least \textbf{18 times} the success rate of the baseline RL policies, even while giving the baseline RL policies access to at least 50\% more simulator steps for training. Our method attains the highest success rate among the ablations for our feedback mechanism. Note: our method and ablations do not all start at 0 wall-clock time, because there is an initial phase of data generation without policy training for each, longer for the No Feedback ablation since it does not get further policy / planner feedback afterward. The ablations attain much stronger performance than the RL baselines, but start to struggle more with parsing language hints. All methods are run with a single seed.}
  \label{fig:results}
\end{figure*}

\begin{figure*}[t]
    \centering
    \includegraphics[width=\linewidth]{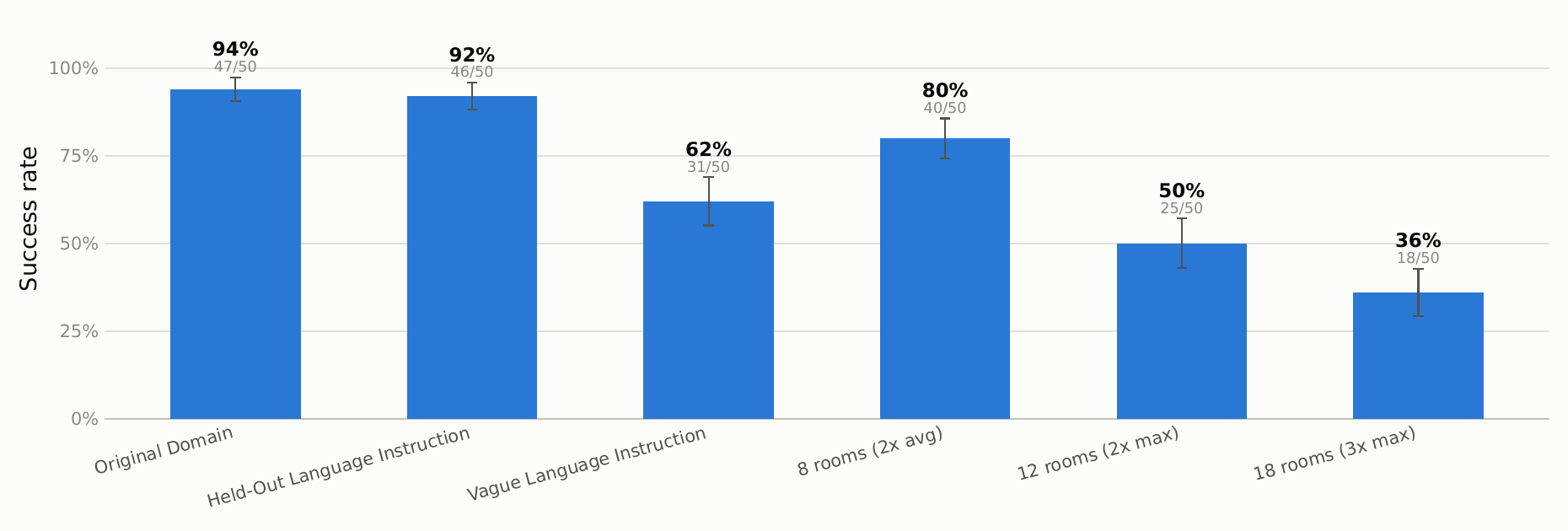}
    \caption{Results of the generalization study. The policy from our method, trained in \autoref{fig:results_lang}, transfers to unseen language commands, to intentionally vague commands, and to problems with vastly more rooms than was seen in training. This shows that our policy learns to comply with physical and epistemic constraints reliably, even in floorplans and given language hints that were unseen in training. Each success rate is calculated based on 50 episodes, and the error bars are binomial standard errors.}
    \label{fig:language-hint-evals}
\end{figure*}

\subsubsection{Baselines}

We compare the performance of our method against two RL baselines: \textbf{Flat RL}, simply training a single policy to solve this task with PPO, and \textbf{Hierarchical RL}, where the agent does RL on which high-level actions to take, given the low-level control policies obtained by distillation for our method. The reward for both baselines was a potential-based reward, rewarding shortened distance to the nearest unseen candidate hint location before the hint is seen and distance to the appropriate marker after the hint was seen. The distance measure used took the geometry of rooms into account (e.g., routed agents through doors to avoid collision) and assigned a large penalty to hitting a wall or to inappropriate guessing actions. This was the best among rewards attempted for the baselines. We used the same encoder backbone and skip connection architecture as in our method and gave these baselines more wall-clock time and training timesteps than our method. We also ran our distillation pipeline with the \textbf{Task Exploration Node} from \cite{mcdonald2022guided}, as a baseline for comparing our feedback mechanism with their high-level task feedback. Due to the simplicity of the control task in Scavenger, we omitted the policy guidance component of their method here. 

\subsubsection{Ablations}

We run three ablations of our two changes. We ran a \textbf{No Feedback} ablation, where the distillation pipeline operates without feedback from the policy of any kind, a \textbf{No Ratio Cap} ablation, where we removed the optimality ratio $\rho$ constraint on successful episodes to be fed into the dataset for reinforcement, and \textbf{No Task-Change}, where we had $\pi_H$ train purely based on imitation loss, i.e. with $\alpha=1, \beta=0$ as described in Dual Averaging, c.f. \autoref{sec:methods_loss}. These were each run for the same number of wall-clock hours as our method, except No Feedback (see below), with possibly differing allocation between data generation and policy training. We found that, since no subsequent policy / planner feedback was provided after the start of training for the No Feedback ablation, it did best when given twice the wall-clock time for generating more initial training data. 

\subsubsection{Architecture and Hyperparameters}

Since the observations consist of two 128$\times$128 images, we train two separate image encoders and then fuse the image latents into one. This is passed to an MLP layer for the $\pi_L$ or to an LSTM with a skip connection for the $\pi_H$. For the language hint, the language command was represented with a SentenceTransformer all-MiniLM-L6-v2 embedding \cite{reimers-2019-sentence-bert}, and concatenated to be fed into LSTM input (for RL, an initial linear layer was added before being concatenated with other LSTM input, for stability). Also, we used $\alpha = \beta = 1$ for the Task-Change Upweighting loss, and $\rho=1.1$ for $\rho$-optimal reinforcement. Flat RL was trained with a batch size of 512 timesteps and constant learning rate of $10^{-4}$, while Hierarchical RL was trained with a batch size of 256 high-level actions and the same constant learning rate. Our method, as well as all ablations, had $\pi_H$ trained on learning rate schedules with warmup, peaking at $3 \times 10^{-4}$ and decaying to $5 \times 10^{-5}$, and a constant $10^{-4}$ learning rate for the $\pi_L$. Further, the task networks had batches of $16$ episodes, and the $\pi_L$ had batches of $64$ timesteps. 

\section{RESULTS}

\subsection{Comparison with Baselines and Ablations}

For both visual and language hints, the distilled planner with $\rho = 1.1$ reinforcement performed best of all methods. Our method attained a success rate above 90\% in both cases. Neither RL variant exceeds 5\% success on either variant of this task, despite being given $\ge 24$ hours of wall-clock compute time and dense reward. Interestingly, when observing rollouts, the flat variant actually does manage to \textit{locate} the hint in both variants nearly 100\% of the time; however, it fails to learn how to use this information and guess the correct marker, despite the reward from that point on densely rewarding consistent progress towards one determined marker. 

\subsection{Generalization Study}

\begin{figure}[t]
    \centering
    \includegraphics[width=0.6\linewidth]{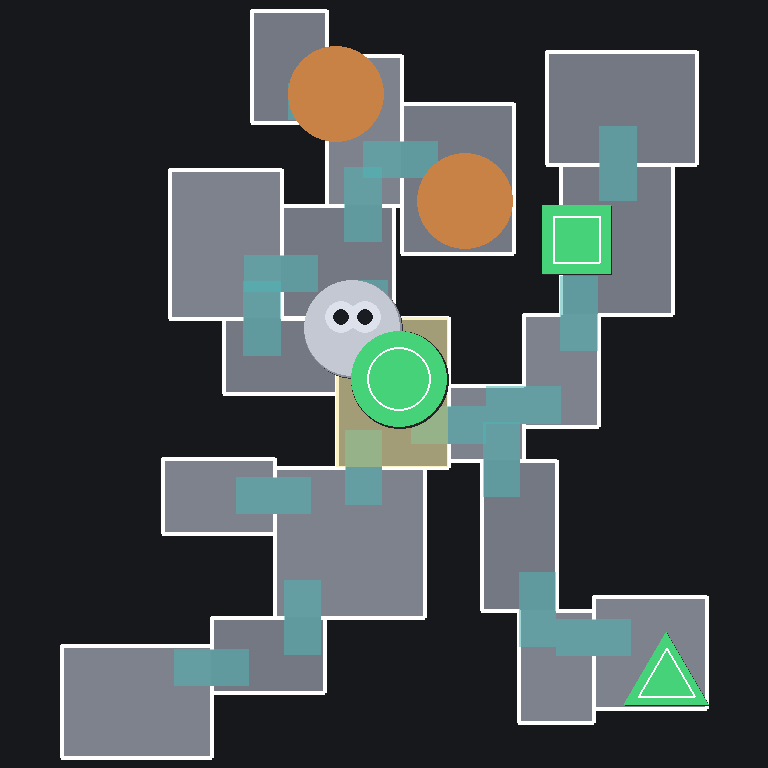}
    \caption{We evaluate transfer from the 3-6 room problems seen in training, to 18 room problems like those shown above, three times the maximum number of rooms seen in training.}
    \label{fig:transfer_rooms_hard_map}
\end{figure}

Since we used learning as a shortcut to avoid the need for belief-state estimation and online planning, we wanted to study how the distilled planner generalized in situations not seen in the training data. To do this, we took the policy that our method trained with language hints (generating the red curve in \autoref{fig:results_lang}). We then ran it for 50 episodes on (a) language commands generated the same way as those seen in training, but held out, and (b) language commands that were designed to be intentionally vague. For example, instead of ``Head to the marker shaped like an upward pointed triangle'' or ``Move toward the upward triangle on the floor,'' both real example commands for the triangle marker seen in training, we evaluate how the agent does on vague commands like ``Navigate to the marker that looks like a capital A without the crossbar'' or ``Make your way to the floor symbol that tapers to one corner at the top'' or ``Approach the floor icon that narrows from a wide base up to one tip.'' 

To test how well the policy generalizes to unseen floorplans, we then ran the model with the same language commands as seen in training, but instead of 3-6 rooms, evaluating (a) with 8 rooms, (b) with 12 rooms (2x the max number of rooms seen in training), and (c) with 18 rooms (3x the max number of rooms seen in training). We find that the trained policy exhibits strong generalization in each of these cases, with virtually no degradation for held-out commands, $>60\%$ success at following indirect descriptions. Moreover, it attained $80\%$ success rate at 8-room problems, $50\%$ for 12-room problems, and $36\%$ success for 18-room problems, three times the maximum seen in training (\autoref{fig:language-hint-evals}). 

\section{DISCUSSION AND FUTURE WORK}

\textbf{Limitations:} Our work assumes that we are able to write a tractable model of the problem for belief-space planning and that the main difficulty is producing a policy that acts on raw observations. This is reasonable in many robotic settings, where access to a simulator allows a system designer to model problems easily, but exploring how to move past this is interesting future work. Further, exploring how to formalize non-robotic safety constraints, e.g., in terminal or computer use, would give us a new programming language for specifying agentic behavior.

In this work, we have contributed to the theory of belief-space planners by characterizing behavior for static search problems, but relaxing these assumptions is future work for the theoretical foundations of robotics. Further, combining imitation learning theory with soundness and completeness guarantees for the planner, would help move towards providing a formal certificate for the distilled planner as well, allowing us to formalize a safety case for a policy trained with this pipeline. Finally, choosing a belief-space planner and using $\rho$-optimal reinforcement both provide \textit{realizable} objectives for the learner: exploring realizable objectives further is a promising algorithmic paradigm for developing agents with epistemic safety cases.

\bibliographystyle{IEEEtran}
\bibliography{ref}

\newpage
\section{SUPPLEMENTAL MATERIALS}

\subsection{TAMP Formalism for Scavenger}\label{section:formalism}

We modeled the Scavenger problem as a 4-action problem: \texttt{nav\_to\_targ\_and\_observe}, \texttt{nav\_to\_other\_room}, and \texttt{nav\_to\_certain\_targ}, and \texttt{validate\_observe}. The last of these was a no-op action that exists only for the sake of the planner and hence was not used for imitation. Below, we outline the domain description used for Scavenger. The predicates in \texttt{nav\_to\_targ\_and\_observe} marked $^{*}$ are asserted symbolically but checked only by the subsequent \texttt{validate\_observe}. If the hint is not at the assumed location, the rollout fails at \texttt{validate\_observe} and the planner replans from there.
\textsf{IsMLTarget} binds \textit{?target} to the most likely particle of the belief. 

\newcommand{\pr}[2]{\mbox{\textsf{#1}(\textit{#2})}}            
\newcommand{\upr}[2]{\mbox{\textsf{#1}(\textit{#2})$^{*}$}}     
\newcommand{\npr}[2]{\mbox{$\lnot$\textsf{#1}(\textit{#2})}}    

\newenvironment{actionschema}[2]{%
  \par\medskip\noindent
  \begin{minipage}{\linewidth}\raggedright
  \setlength{\parindent}{0pt}%
  \newcommand{\field}[1]{\par\hangindent 5em\hangafter 1
    \makebox[5em][l]{\textbf{##1}}}%
  \textbf{\textsc{#1}}\,(\textit{#2})\par
  \vspace{2pt}\hrule\vspace{4pt}%
}{\par\end{minipage}\par\medskip}

\begin{actionschema}{nav\_to\_targ\_and\_observe}{?robot - Robot, ?target - Target, ?room - Room}
  \field{pre}
    \pr{RobotInRoom}{?robot, ?room},\ \
    \pr{TargetInRoom}{?target, ?room},\ \
    \pr{IsMLTarget}{?robot, ?target} 
  \field{eff}
    \pr{RobotAtTarget}{?robot, ?target},\ \
    \upr{CertainHintTargLoc}{?target},\ \
    \upr{ScavengerTargetKnown}{?target},\ \
    \pr{RobotInRoom}{?robot, ?room}
  \field{sampler}
    \textit{?robot} $\leftarrow$ \textsf{nav\_2d\_pose\_sampler}
\end{actionschema}

\begin{actionschema}{validate\_observe}{?target - Target, ?robot - Robot, ?room - Room}
  \field{pre}
    \pr{CertainHintTargLoc}{?target},\ \
    \pr{ScavengerTargetKnown}{?target},\ \
    \pr{TargetInRoom}{?target, ?room},\ \
    \pr{RobotInRoom}{?robot, ?room}
  \field{eff}
    \pr{SuccessfulObserve}{?target},\ \
    \pr{ReadyToNavigate}{?robot}
  \field{sampler}
    \textit{?robot} $\leftarrow$ \textsf{nav\_2d\_pose\_sampler}
\end{actionschema}

\begin{actionschema}{nav\_to\_certain\_targ}{?robot - Robot, ?target - Target, ?room - Room}
  \field{pre}
    \pr{ScavengerTargetKnown}{?target},\ \
    \pr{SuccessfulObserve}{?target},\ \
    \pr{RobotInRoom}{?robot, ?room},\ \
    \pr{ScavengerTargetInRoom}{?target, ?room}
  \field{eff}
    \pr{RobotAtCertainTarget}{?robot, ?target},\ \
    \pr{GuessAction}{?robot}
      \field{sampler}
    \textit{?robot} $\leftarrow$ \textsf{nav\_2d\_pose\_sampler}
\end{actionschema}

\begin{actionschema}{nav\_to\_other\_room}{?robot - Robot, ?room - Room, ?room$'$ - Room, ?door - Door, ?v$'$ - Vantage}
  \field{pre}
    \pr{RobotInRoom}{?robot, ?room},\ \
    \pr{DoorConnectsRooms}{?door, ?room, ?room$'$},\ \
    \pr{RoomContainsVantage}{?room, ?v$'$}
  \field{eff}
    $\lnot$ \pr{RobotInRoom}{?robot, ?room},\ \
    \pr{RobotInRoom}{?robot, ?room$'$},\ \
    \pr{RobotAtVantage}{?robot, ?v$'$}
    \field{sampler}
    \textit{?robot} $\leftarrow$ \textsf{nav\_2d\_pose\_sampler}
\end{actionschema}

\end{document}